\def\year{2018}\relax
\documentclass[letterpaper]{article} 
\usepackage{aaai18}  
\usepackage{times}  
\usepackage{helvet}  
\usepackage{courier}  
\usepackage{url}  
\usepackage{graphicx}  
\usepackage{subfigure} 
\usepackage{amsmath,amsfonts,amssymb}
\usepackage{bbm}
\usepackage{color}

\usepackage{algorithm}
\usepackage{algorithmic}
\usepackage{wrapfig}
\usepackage{caption}

\newcommand{\pol}[0]{\boldsymbol{\pi}}
\newcommand{\eps}[0]{\boldsymbol{\varepsilon}}

\newcommand{\insertfigure}[3]{
\begin{figure}[ht]
\centering
\includegraphics[width=#2\textwidth]{figures/#1}
\caption{#3}
\label{fig:#1}
\end{figure}
}
\newcommand{\inserttable}[4]{
\begin{table}[h!]
\centering
\begin{tabular}{#2}
#3
\end{tabular}
\caption{#4}
\label{table:#1}
\end{table}
}

\begin{document}
%
\title{Emergence of Grounded Compositional Language in Multi-Agent Populations}

\author{Igor Mordatch\\ OpenAI \\ San Francisco, California, USA
  \And Pieter Abbeel\\
  UC Berkeley \\ Berkeley, California, USA}

\maketitle
\begin{abstract}
By capturing statistical patterns in large corpora, machine learning has enabled significant advances in natural language processing, including in machine translation, question answering, and sentiment analysis.  However, for agents to intelligently interact with humans, simply capturing the statistical patterns is insufficient. In this paper we investigate if, and how, grounded compositional language can emerge as a means to achieve goals in multi-agent populations.  Towards this end, we propose a multi-agent learning environment and learning methods that bring about emergence of a basic compositional language. This language is represented as streams of abstract discrete symbols uttered by agents over time, but nonetheless has a coherent structure that possesses a defined vocabulary and syntax. We also observe emergence of non-verbal communication such as pointing and guiding when language communication is unavailable. 
\end{abstract}

\section{Introduction}
\label{sec:introduction}

Development of agents that are capable of communication and flexible language use is one of the long-standing challenges facing the field of artificial intelligence. Agents need to develop communication if they are to successfully coordinate as a collective. Furthermore, agents will need some language capacity if they are to interact and productively collaborate with humans or make decisions that are interpretable by humans. If such a capacity were to arise artificially, it could also offer important insights into questions surrounding development of human language and cognition.

But if we wish to arrive at formation of communication from first principles, it must form out of necessity. The approaches that learn to plausibly imitate language from examples of human language, while tremendously useful, do not learn \emph{why} language exists. Such supervised approaches can capture structural and statistical relationships in language, but they do not capture its functional aspects, or that language happens for purposes of successful coordination between humans. Evaluating success of such imitation-based approaches on the basis of linguistic plausibility also presents challenges of ambiguity and requirement of human involvement.

Recently there has been a surge of renewed interest in the pragmatic aspects of language use and it is also the focus of our work. We adopt a view of \cite{mordatch16} that an agent possesses an understanding of language when it can use language (along with other tools such as non-verbal communication or physical acts) to accomplish goals in its environment. This leads to evaluation criteria that can be measured precisely and without human involvement. 

In this paper, we propose a physically-situated multi-agent learning environment and learning methods that bring about emergence of a basic compositional language. This language is represented as streams of abstract discrete symbols uttered by agents over time, but nonetheless has a coherent structure that possesses a defined vocabulary and syntax. The agents utter communication symbols alongside performing actions in the physical environment to cooperatively accomplish goals defined by a joint reward function shared between all agents. There are no pre-designed meanings associated with the uttered symbols - the agents form concepts relevant to the task and environment and assign arbitrary symbols to communicate them.

There are similarly no explicit language usage goals, such as making correct utterances, and no explicit roles agents are assigned, such as speaker or listener, or explicit turn-taking dialogue structure as in traditional language games. There may be an arbitrary number of agents in a population communicating at the same time and part of the difficulty is learning to refer specific agents. A population of agents is situated as moving particles in a continuous two-dimensional environment, possessing properties such as color and shape. The goals of the population are based on non-linguistic objectives, such as moving to a location and language arises from the need to coordinate on those goals. We do not rely on any supervision such as human demonstrations or text corpora. 

Similar to recent work,
we formulate the discovery the action and communication protocols for our agents jointly as a reinforcement learning problem. Agents perform physical actions and communication utterances according to an identical policy that is instantiated for all agents and fully determines the action and communication protocols. The policies are based on neural network models with an architecture composed of dynamically-instantiated recurrent modules. This allows decentralized execution with a variable number of agents and communication streams. The joint dynamics of all agents and environment, including discrete communication streams are fully-differentiable, the agents' policy is trained end-to-end with backpropagation through time.

The languages formed exhibit interpretable compositional structure that in general assigns symbols to separately refer to environment landmarks, action verbs, and agents. However, environment variation leads to a number of specialized languages, omitting words that are clear from context. For example, when there is only one type of action to take or one landmark to go to, words for those concepts do not form in the language. Considerations of the physical environment also have an impact on language structure. For example, a symbol denoting \emph{go} action is typically uttered first because the listener can start moving before even hearing the destination. This effect only arises when linguistic and physical behaviors are treated jointly and not in isolation.

The presence of a physical environment also allows for alternative strategies aside from language use to accomplish goals. A visual sensory modality provides an alternative medium for communication and we observe emergence of non-verbal communication such as pointing and guiding when language communication is unavailable. When even non-verbal communication is unavailable, strategies such as direct pushing may be employed to succeed at the task. It is important to us to build an environment with a diverse set of capabilities which language use develops alongside with.


By compositionality we mean the combination of multiple words to create meaning, as opposed to holistic languages that have a unique word for every possible meaning \cite{kirby01spontaneous}. Our work offers insights into why such compositional structure emerges. In part, we find it to emerge when we explicitly encourage active vocabulary sizes to be small through a soft penalty. This is consistent with analysis in evolutionary linguistics \cite{nowak00syntax} that finds composition to emerge only when number of concepts to be expressed becomes greater than a factor of agent's symbol vocabulary capacity. Another important component leading to composition is training on a variety of tasks and environment configurations simultaneously. Training on cases where most information is clear from context (such as when there is only one landmark) leads to formation of atomic concepts that are reused compositionally in more complicated cases.


\section{Related Work}
\label{sec:related}


Recent years have seen substantial progress in practical natural language applications such as machine translation \cite{sutskever14,bahdanau14}, sentiment analysis \cite{socher13}, document summarization \cite{durrett16}, and domain-specific dialogue \cite{dhingra16}. Much of this success is a result of intelligently designed statistical models trained on large static datasets. However, such approaches do not produce an understanding of language that can lead to productive cooperation with humans.




An interest in pragmatic view of language understanding has been longstanding \cite{austin62,grice75} and has recently argued for in \cite{mordatch16,lake16,lazaridou16}. Pragmatic language use has been proposed in the context of two-player reference games \cite{klein10,vogel14,andreas16} focusing on the task of identifying object references through a learned language. \cite{winograd73,wang16} ground language in a physical environment and focusing on language interaction with humans for completion of tasks in the physical environment. In such a pragmatic setting, language use for communication of spatial concepts has received particular attention in \cite{steels95,spatial16}.

Aside from producing agents that can interact with humans through language, research in pragmatic language understanding can be informative to the fields of linguistics and cognitive science. Of particular interest in these fields has been the question of how syntax and compositional structure in language emerged, and why it is largely unique to human languages \cite{kirby_syntax99,nowak00syntax,steels05syntax}. Models such as Rational Speech Acts \cite{rsa12} and Iterated Learning \cite{kirby14iterated} have been popular in cognitive science and evolutionary linguistics, but such approaches tend to rely on pre-specified procedures or models that limit their generality.

The recent work that is most similar to ours is the application of reinforcement learning approaches towards the purposes of learning a communication protocol, as exemplified by \cite{bratman10,foerster16b,fergus16,lazaridou16b}.










\section{Problem Formulation}
The setting we are considering is a cooperative partially observable Markov game \cite{littman94}, which is a multi-agent extension of a Markov decision process. A Markov game for $N$ agents is defined by set of states $\mathcal{S}$ describing the possible configurations of all agents, a set of actions $\mathcal{A}_1,...,\mathcal{A}_N$ and a set of observations $\mathcal{O}_1,...,\mathcal{O}_N$ for each agent. Initial states are determined by a distribution $\rho : \mathcal{S} \mapsto [0,1]$. State transitions are determined by a function $\mathcal{T} : \mathcal{S} \times \mathcal{A}_1 \times ... \times \mathcal{A}_N \mapsto \mathcal{S}$.
For each agent $i$, rewards are given by function
$r_i : \mathcal{S} \times \mathcal{A}_i \mapsto \mathbb{R}$, 
observations are given by function $\mathbf{o}_i : \mathcal{S} \mapsto \mathcal{O}_i$.
To choose actions, each agent $i$ uses a stochastic policy $\pol_i : \mathcal{O}_i \times \mathcal{A}_i \mapsto [0,1]$.

In this work, we assume all agents have identical action and observation spaces, and all agents act according to the same policy $\pol$ and receive a shared reward. We consider a finite horizon setting, with episode length $T$. In a cooperative setting, the problem is to find a policy that maximizes the expected shared return for all agents, which can be solved as a joint minimization problem:
\begin{equation*}
    \max_{\pol} R(\pol), \quad \text{where} \quad
    R(\pol) = \mathbb{E}\bigg[ \sum_{t=0}^T \sum_{i=0}^N r(\mathbf{s}^t_i, \mathbf{a}^t_i) \bigg]
\end{equation*}

\section{Grounded Communication Environment}
\label{sec:environment}

\insertfigure{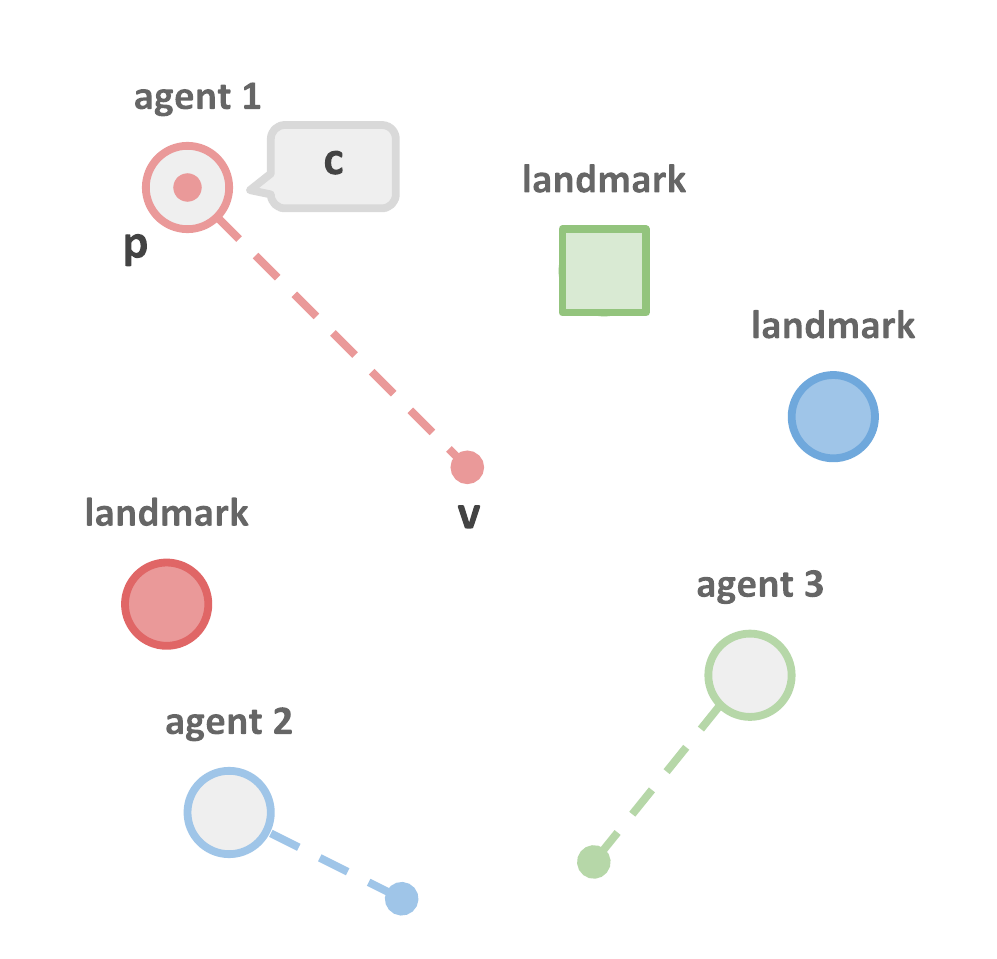}{0.35}{An example of environments we consider.
}

As argued in the introduction, grounding multi-agent communication in a physical environment is crucial for interesting communication behaviors to emerge. In this work, we consider a physically-simulated two-dimensional environment in continuous space and discrete time. This environment consists of $N$ agents and $M$ landmarks. Both agent and landmark entities inhabit a physical location in space $\mathbf{p}$ and posses descriptive physical characteristics, such as color and shape type. In addition, agents can direct their gaze to a location $\mathbf{v}$.
Agents can act to move in the environment and direct their gaze, but may also be affected by physical interactions with other agents. We denote the physical state of an entity (including descriptive characteristics) by $\mathbf{x}$ and describe its precise details and transition dynamics in the Appendix.


In addition to performing physical actions, agents utter verbal communication symbols $c$ at every timestep. These utterances are discrete elements of an abstract symbol vocabulary $\mathcal{C}$ of size $K$. We do not assign any significance or meaning to these symbols. They are treated as abstract categorical variables that are emitted by each agent and observed by all other agents. It is up to agents at training time to assign meaning to these symbols. As shown in Section \ref{sec:experiments}, these symbols become assigned to interpretable concepts. Agents may also choose not to utter anything at a given timestep, and there is a cost to making an utterance, loosely representing the metabolic effort of vocalization. We denote a vector representing one-hot encoding of symbol $c$ with boldface $\mathbf{c}$.

Each agent has internal goals specified by vector $\mathbf{g}$ that are private and not observed by other agents. These goals are grounded in the physical environment and include tasks such as moving to or gazing at a location. These goals may involve other agents (requiring the other agent to move to a location, for example) but are not observed by them and thus necessitate coordination and communication between agents. Verbal utterances are one tool which the agents can use to cooperatively accomplish all goals, but we also observe emergent use of non-verbal signals and altogether non-communicative strategies.

To aid in accomplishing goals, each agent has internal recurrent memory bank $\mathbf{m}$ that is also private and not observed by other agents. This memory bank has no pre-designed behavior and it is up to the agents to learn to utilize it appropriately. 

The full state of the environment is given by
$
    \mathbf{s} = \begin{bmatrix} \ \mathbf{x}_{1,...,(N+M)} \ \mathbf{c}_{1,...,N} \ \mathbf{m}_{1,...,N} \ \mathbf{g}_{1,...,N} \ \end{bmatrix} \in \mathcal{S}.
$
Each agent observes physical states of all entities in the environment, verbal utterances of all agents, and its own private memory and goal vector. The observation for agent $i$ is
$
    \mathbf{o}_i(\mathbf{s}) = \begin{bmatrix} \ _i\mathbf{x}_{1,...,(N+M)} \ \mathbf{c}_{1,...,N} \ \mathbf{m}_i \ \mathbf{g}_i \ \end{bmatrix}.
$
Where $_i\mathbf{x}_j$ is the observation of entity $j$'s physical state in agent $i$'s reference frame (see Appendix for details).
More intricate observation models are possible, such as physical observations solely from pixels or verbal observations from a single input channel. These models would require agents learning to perform visual processing and source separation, which are orthogonal to this work. Despite the dimensionality of observations varying with the number of physical entities and communication streams, our policy architecture as described in Section \ref{sec:policy} allows a single policy parameterization across these variations.

\insertfigure{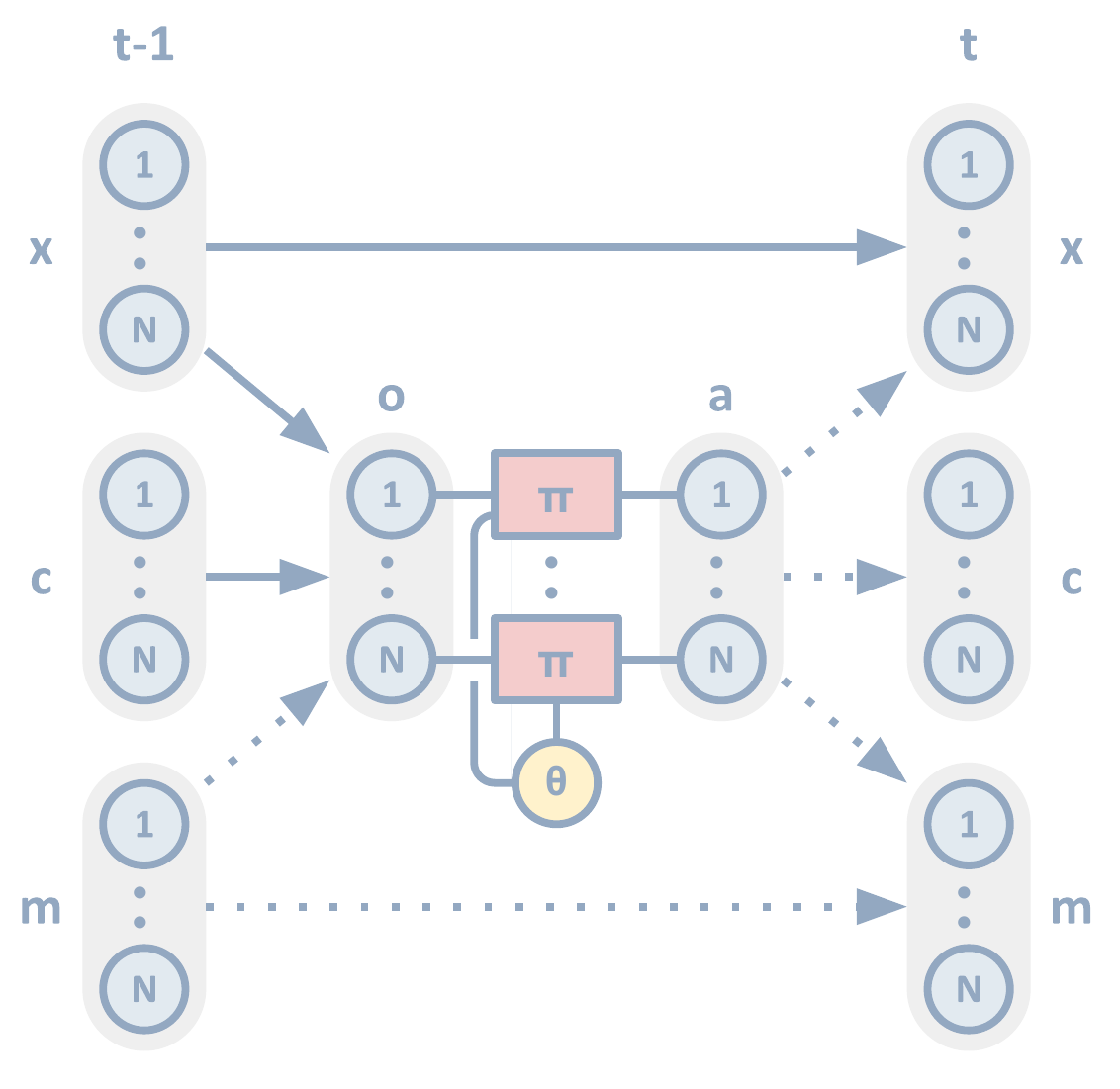}{0.40}{The transition dynamics of $N$ agents from time $t-1$ to $t$. Dashed lines indicate one-to-one dependencies between agents and solid lines indicate all-to-all dependencies.}

\section{Policy Learning with Backpropagation}
Each agent acts by sampling actions from a stochastic policy $\pol$, which is identical for all agents and defined by parameters $\theta$. There are several common options for finding optimal policy parameters. The model-free framework of Q-learning can be used to find the optimal state-action value function, and employ a policy that acts greedily to according to the value function. Unfortunately, Q function dimensionality scales quadratically with communication vocabulary size, which can quickly become intractably large. Alternatively it is possible to directly learn a policy function using model-free policy gradient methods, which use sampling to estimate the gradient of policy return $\frac{dR}{d\theta}$. The gradient estimates from these methods can exhibit very high variance and credit assignment becomes an especially difficult problem in the presence of sequential communication actions.

Instead of using model-free reinforcement learning methods, we build an end-to-end differentiable model of all agent and environment state dynamics over time and calculate $\frac{dR}{d\theta}$ with backpropagation. At every optimization iteration, we sample a new batch of 1024 random environment instantiations and backpropagate their dynamics through time to calculate the total return gradient. Figure \ref{fig:timestep} shows the dependency chain between two timesteps. A similar approach was employed by \cite{foerster16b,fergus16} to compute gradients for communication actions, although the latter still employed model-free methods for physical action computation.


The physical state dynamics, including discontinuous contact events can be made differentiable with smoothing. However, communication actions require emission of discrete symbols, which present difficulties for backpropagation.

\subsection{Discrete Communication and Gumbel-Softmax Estimator}
\label{sec:gumbel}
In order to use categorical communication emissions $\mathbf{c}$ in our setting, it must be possible to differentiate through them. There has been a wealth of work in machine learning on differentiable models with discrete variables, but we found recent approach in \cite{gumbel16a,gumbel16b} to be particularly effective in our setting. The approach proposes a Gumbel-Softmax distribution, which is a continuous relaxation of a discrete categorical distribution. Given $K$-categorical distribution parameters $p$, a differentiable $K$-dimensional one-hot encoding sample $G$ from the Gumbel-Softmax distribution can be calculated as:
\begin{align*}
G(\text{log}p)_k = \frac{\text{exp}((\text{log}p_k + \eps)/\tau )}{\sum_{j=0}^K \text{exp}((\text{log}p_j + \eps)/\tau ) }
\end{align*}
Where $\eps$ are i.i.d. samples from Gumbel$(0,1)$ distribution,
$\eps = -\text{log}(-\text{log}(u)), \; \; u \sim \mathcal{U}[0,1]$
and $\tau$ is a softmax temperature parameter. We did not find it necessary to anneal the temperature and set it to $1$ in all our experiments for training and sample directly from the categorical distribution at test time. To emit a communication symbol, our policy is trained to directly output $\text{log}p \in \mathcal{R}^K$, which is transformed to a symbol emission sample $\mathbf{c} \sim G(\text{log}p)$. The resulting gradient can be estimated as $\frac{d\mathbf{c}}{d\theta} = \frac{dG}{dp} \frac{dp}{d\theta}$.


\subsection{Policy Architecture}
\label{sec:policy}
\insertfigure{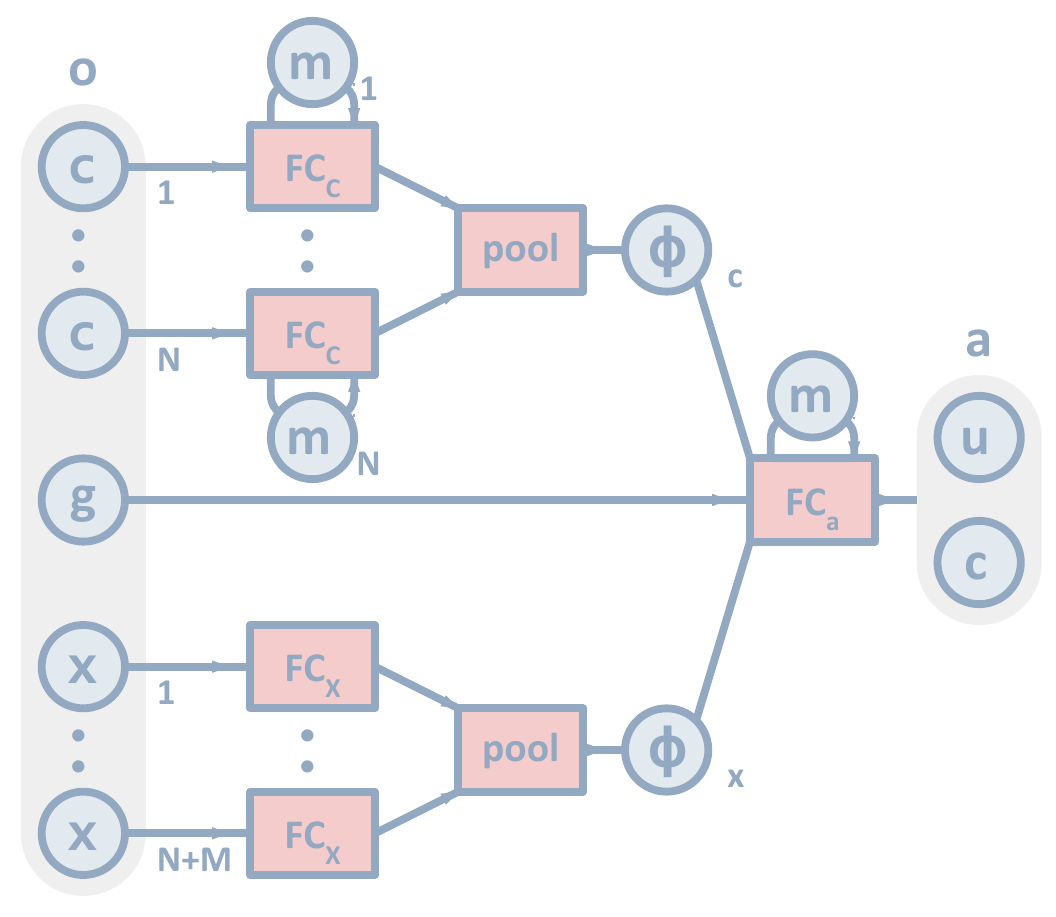}{0.40}{Overview of our policy architecture, mapping observations to actions at every point time time. \emph{FC} indicates a fully-connected processing module that shares weights with all others of its label. \emph{pool} indicates a softmax pooling layer.}
The policy class we consider in this work are stochastic neural networks. The policy outputs samples of an agent's physical actions $\mathbf{u}$, communication symbol utterance $\mathbf{c}$, and internal memory updates $\Delta \mathbf{m}$. The policy must consolidate multiple incoming communication symbol streams emitted by other agents, as well as incoming observations of physical entities. Importantly, the number of agents (and thus the number of communication streams) and number of physical entities can vary between environment instantiations. To support this, the policy instantiates a collection of identical processing modules for each communication stream and each observed physical entity. Each processing module is a fully-connected multi-layer perceptron. The weights between all communication processing and physical observation modules are shared. The outputs of individual processing modules are pooled with a softmax operation into feature vectors $\boldsymbol{\phi}_c$ and $\boldsymbol{\phi}_x$ for communication and physical observation streams, respectively. Such weight sharing and pooling makes it possible to apply the same policy parameters to any number of communication and physical observations. 

The pooled features and agent's private goal vector are passed to the final processing module that outputs distribution parameters $\begin{bmatrix} \ \boldsymbol{\psi}_u \ \boldsymbol{\psi}_c \ \end{bmatrix}$ from which action samples are generated as $\mathbf{u} = \boldsymbol{\psi}_u + \eps$ and $\mathbf{c} \sim G(\boldsymbol{\psi}_c)$, where $\eps$ is a zero-mean Gaussian noise.

Unlike communication games where agents only emit a single utterance, our agents continually emit a stream of symbols over time. Thus processing modules that read and write communication utterance streams benefit greatly from recurrent memory that can capture meaning of a stream over time. To this end, we augment each communication processing and output module with an independent internal memory state $\mathbf{m}$, and each module outputs memory state updates $\Delta \mathbf{m}$. In this work we use simple additive memory updates $\mathbf{m}^t = \text{tanh}(\mathbf{m}^{t-1} + \Delta \mathbf{m}^{t-1} + \eps)$ for simplicity and interpretability, but other memory architectures such LSTMs can be used.
We build all fully-connected modules with 256 hidden units and 2 layers each in all our experiments, using exponential-linear units and dropout with a rate of 0.1 between all hidden layers. Size is feature vectors $\boldsymbol{\phi}$ is 256 and size of each memory module is 32. The overall policy architecture is shown in Figure \ref{fig:policy_noarrow}.


\subsection{Auxiliary Prediction Reward}
To help policy training avoid local minima in more complex environments, we found it helpful to include auxiliary goal prediction tasks, similar to recent work in reinforcement learning \cite{dosovitskiy16,silver16}. In agent $i$'s policy, each communication processing module $j$ additionally outputs a prediction $\hat{\mathbf{g}}_{i,j}$ of agent $j$'s goals. We do not use $\hat{\mathbf{g}}$ as an input in calculating actions. It is only used for the purposes of auxiliary prediction task. At the end of the episode, we add a reward for predicting other agent's goals, which in turn encourages communication utterances that convey the agent's goals clearly to other agents. Across all agents this reward has the form:
\begin{align*}
    r_g = -\sum_{\left\{i,j | i \neq j \right\}} \| \hat{\mathbf{g}}^T_{i,j} - \mathbf{g}^T_j \|^2
\end{align*}

\section{Compositionality and Vocabulary Size}
What leads to compositional syntax formation? One known constructive hypothesis requires modeling the process of language transmission and acquisition from one generation of agents to the next iteratively as in \cite{kirby14iterated}. In such iterated learning setting, compositionality emerges due to poverty of stimulus - one generation will only observe a limited number of symbol utterances from the previous generation and must infer meaning of unseen symbols. This approach requires modeling language acquisition between agents, but when implemented with pre-designed rules was shown over multiple iterations between generations to lead to formation of a compositional vocabulary.

Alternatively, \cite{nowak00syntax} observed that emergence of compositionality requires the number of concepts describable by a language to be above a factor of vocabulary size. In our preliminary environments the number of concepts to communicate is still fairly small and is within the capacity of a non-compositional language. We use a maximum vocabulary size $K=20$ in all our experiments. We tested a smaller maximum vocabulary size, but found that policy optimization became stuck in a poor local minima where concepts became conflated. Instead, we propose to use a large vocabulary size limit but use a soft penalty function to prevent the formation of unnecessarily large vocabularies. This allows the intermediate stages of policy optimization to explore large vocabularies, but then converge on an appropriate active vocabulary size. As shown in Figure \ref{fig:wordcount}, this is indeed what happens.

How do we penalize large vocabulary sizes?  \cite{nowak00syntax} proposed a word population dynamics model that defines reproductive ratios of words to be proportional to their frequency, making already popular words more likely to survive. Inspired by these rich-get-richer dynamics, we model the communication symbols as being generated from a Dirichlet Process \cite{teh11dirichlet}. Each communication symbol has a probability of being symbol $c_k$ as
\begin{align*}
    p(c_k) = \frac{ n_k }{\alpha + n - 1}
\end{align*}
Where $n_k$ is the number of times symbol $c_k$ has been uttered and $n$ is the total number of symbols uttered. These counts are accumulated over agents, timesteps, and batch entries. $\alpha$ is a Dirichlet Process hyperparameter corresponding to the probability of observing an out-of-vocabulary word. The resulting reward across all agents is the log-likelihood of all communication utterances to independently have been generated by a Dirichlet Process:
\begin{align*}
    r_c = \sum_{i,t,k} \mathbbm{1}[\mathbf{c}^t_i = c_k] \text{log} p(c_k)
\end{align*}
Maximizing this reward leads to consolidation of symbols and the formation of compositionality. This approach is similar to encouraging code population sparsity in autoencoders \cite{ng11}, which was shown to give rise to compositional representations for images.

\section{Experiments}
\label{sec:experiments}

We experimentally investigate how variation in goals, environment configuration, and agents’ physical capabilities lead to different communication strategies. In this work, we consider three types of actions an agent needs to perform: \emph{go to} location, \emph{look at} location, and \emph{do nothing}. Goal for agent $i$ consists of an action to perform, a location to perform it on $\bar{\mathbf{r}}$, and an agent $r$ that should perform that action. These goal properties are accumulated into goal description vector $\mathbf{g}$. These goals are private to each agent, but may involve other agents. For example, agent $i$ may want agent $r$ to go to location $\bar{\mathbf{r}}$. This goal is not observed by agent $r$, and requires communication between agents $i$ and $r$. The goals are assigned to agents such that no agent receives conflicting goals. We do however show generalization in the presence of conflicting goals in Section \ref{sec:generalization}.

Agents can only communicate in discrete symbols and have individual reference frames without a shared global positioning reference (see Appendix), so cannot directly send goal position vector. What makes the task possible is that we place goal locations $\bar{\mathbf{r}}$ on landmark locations of which are observed by all agents (in their invidiaul reference frames). The strategy then is for agent $i$ to unambiguously communicate landmark reference to agent $r$. Importantly, we do not provide explicit association between goal positions and landmark reference. It is up to the agents to learn to associate a position vector with a set of landmark properties and communicate them with discrete symbols.

In the results that follow, agents do not observe other agents. This disallows capacity for non-verbal communication, necessitating the use of language. In section \ref{sec:nonverbal} we report what happens when agents are able to observe each other and capacity for non-verbal communication is available.

Despite training with continuous relaxation of the categorical distribution, we observe very similar reward performance at test time. No communication is provided as a baseline (again, non-verbal communication is not possible). The no-communication strategy is for all agents go towards the centroid of all landmarks.
\inserttable{reward}{|c|c|c|}{
\hline
\textbf{Condition} & \textbf{Train Reward} & \textbf{Test Reward} \\
\hline
No Communication & -0.919 & -0.920 \\
\hline
Communication & -0.332 & -0.392 \\
\hline
}{Training and test physical reward for setting with and without communication.}

\subsection{Syntactic Structure}
\insertfigure{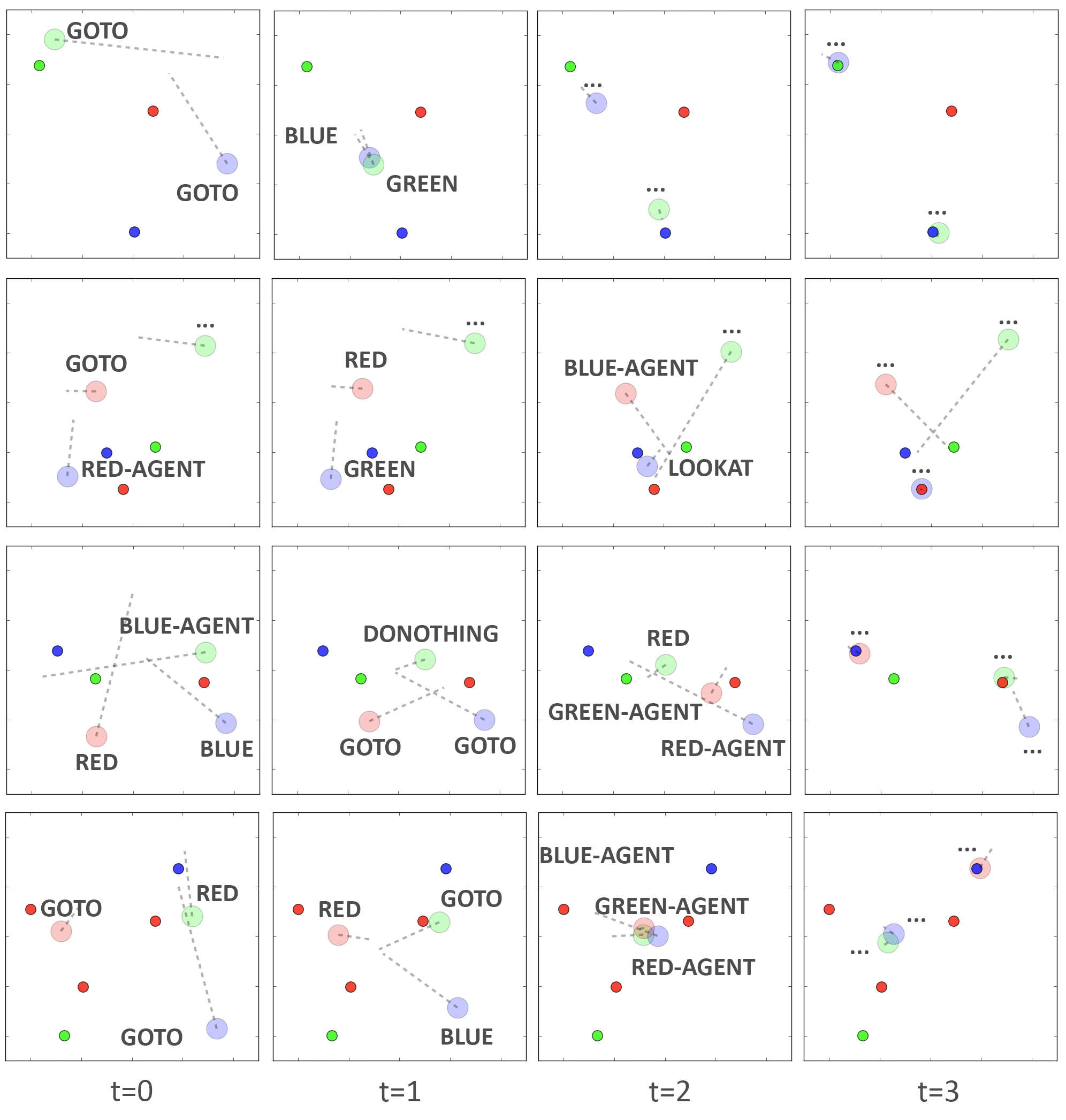}{0.45}{A collection of typical sequences of events in our environments shown over time. Each row is an independent trial. Large circles represent agents and small circles represent landmarks. Communication symbols are shown next to the agent making the utterance. The labels for abstract communication symbols are chosen purely for visualization and \emph{...} represents silence symbol.}
We observe a compositional syntactic structure emerging in the stream of symbol uttered by agents. When trained on environments with only two agents, but multiple landmarks and actions, we observe symbols forming for each of the landmark colors and each of the action types. A typical conversation and physical agent configuration is shown in first row of Figure \ref{fig:env_streams} and is as follows:
\begingroup
\fontsize{8pt}{8pt}\selectfont
\begin{verbatim}
  Green Agent: GOTO, GREEN, ...
  Blue Agent: GOTO, BLUE,  ...
\end{verbatim}
\endgroup
The labels for abstract symbols are chosen by us purely for interpretability and visualization and carry no meaning for training. While there is recent work on interpreting continuous machine languages \cite{andreas17neuralese}, the discrete nature and small size of our symbol vocabulary makes it possible to manually labels to the symbols. See results in supplementary video for consistency of the vocabulary usage.

Physical environment considerations play a part in the syntactic structure. The action type verb \emph{GOTO} is uttered first because actions take time to accomplish in the grounded environment. When the agent receives \emph{GOTO} symbol it starts moving toward the centroid of all the landmarks (to be equidistant from all of them) and then moves towards the specific landmark when it receives its color identity.

When the environment configuration can contain more than three agents, agents need to form symbols for referring to each other. Three new symbols form to refer to agent colors that are separate in meaning from landmark colors. The typical conversations are shown in second and third rows of Figure \ref{fig:env_streams}.
\begingroup
\fontsize{8pt}{8pt}\selectfont
\begin{verbatim}
  Red Agent: GOTO, RED, BLUE-AGENT, ...
  Green Agent: ..., ..., ..., ...
  Blue Agent: RED-AGENT, GREEN, LOOKAT, ...    
\end{verbatim}
\endgroup
Agents may not omit any utterances when they are the subject of their private goal, in which case they have access to that information and have no need to announce it. In this language, there is no set ordering to word utterances. Each symbol contributes to sentence meaning independently, similar to case marking grammatical strategies used in many human languages \cite{steels_marker15}. 

The agents largely settle on using a consistent set of symbols for each meaning, due to vocabulary size penalties and that discourage synonyms. We show the aggregate streams of communication utterances in Figure \ref{fig:symbol_streams}.
\insertfigure{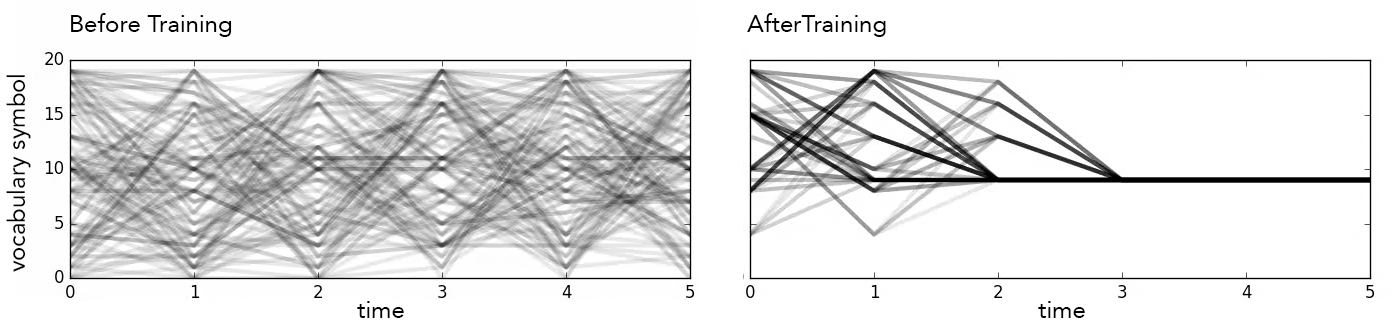}{0.45}{Communication symbol streams emitted by agents over time before and after training accumulated over 10 thousand test trials.}

In simplified environment configurations when there is only one landmark or one type of action to take, no symbols are formed to refer to those concepts because they are clear from context.

\subsection{Symbol Vocabulary Usage}
\insertfigure{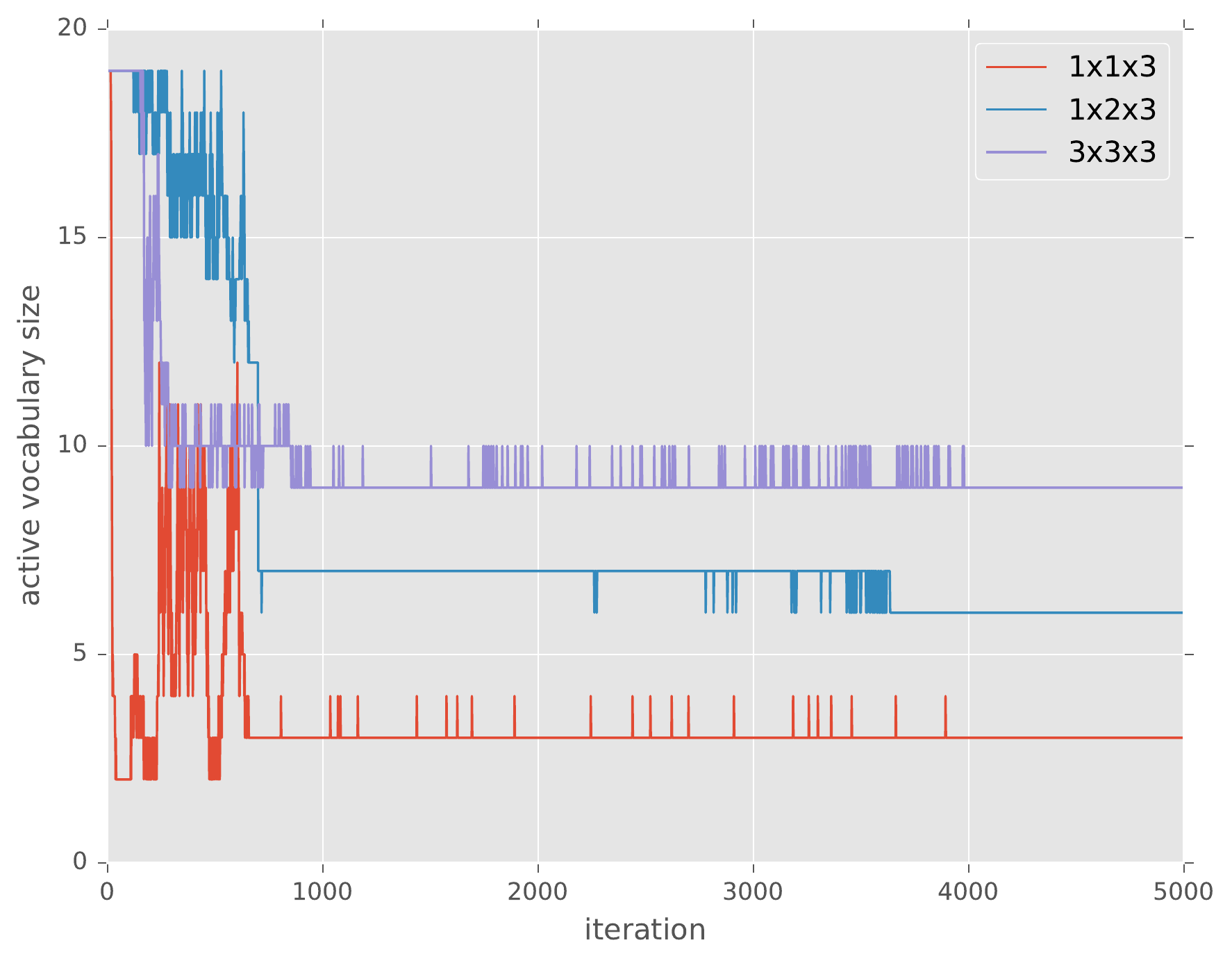}{0.45}{Word activations counts for different environment configurations over training iterations.}
We find word activation counts to settle on the appropriate compositional word counts. That early during training large vocabulary sizes are being taken advantage of to explore the space of communication possibilities before settling on the appropriate effective vocabulary sizes as shown in Figure \ref{fig:wordcount}. In this figure, \emph{1x1x3} case refers to environment with two agents and a single action, which requires only communicating one of three landmark identities. \emph{1x2x3} contains two types of actions, and \emph{3x3x3} case contains three agents that require explicit referencing.


\subsection{Generalization to Unseen Configurations}
\label{sec:generalization}
One of the advantages of decentralised execution policies is that trained agents can be placed into arbitrarily-sized groups and still function reasonably. When there are additional agents in the environment with the same color identity, all agents of the same color will perform the same task if they are being referred to. Additionally, when agents of a particular color are asked to perform two conflicting tasks (such as being asked go to two different landmarks by two different agents), they will perform the average of the conflicting goals assigned to them. Such cases occur despite never having been seen during training.

Due to the modularized observation architecture, the number of landmarks in the environment can also vary between training and execution. The agents perform sensible behaviors with different numbers of landmarks, despite not being trained in such environments. For example, when there are distractor landmarks of novel colors, the agents never go towards them. When there are multiple landmarks of the same color, the agent communicating the goal still utters landmark color (because the goal is the position of one of the landmarks). However, the agents receiving the landmark color utterance go towards the centroid of all landmark of the same color, showing a very sensible generalization strategy. An example of such case is shown in fourth row of Figure \ref{fig:env_streams}.

\subsection{Non-verbal Communication and Other Strategies}
\label{sec:nonverbal}
The presence of a physical environment also allows for alternative strategies aside from language use to accomplish goals. In this set of experiments we enable agents to observe other agents' position and gaze location, and in turn disable communication capability via symbol utterances. When agents can observe each other's gaze, a pointing strategy forms where the agent can communicate a landmark location by gazing in its direction, which the recipient correctly interprets and moves towards. When gazes of other agents cannot be observed, we see behavior of goal sender agent moving towards the location assigned to goal recipient agent (despite receiving no explicit reward for doing so), in order to guide the goal recipient to that location. Lastly, when neither visual not verbal observation is available on part of the goal recipient, we observe the behavior of goal sender directly pushing the recipient to the target location. Examples of such strategies are shown in Figure \ref{fig:non_verbal} and supplementary video. It is important to us to build an environment with a diverse set of capabilities which language use develops alongside with. 
\insertfigure{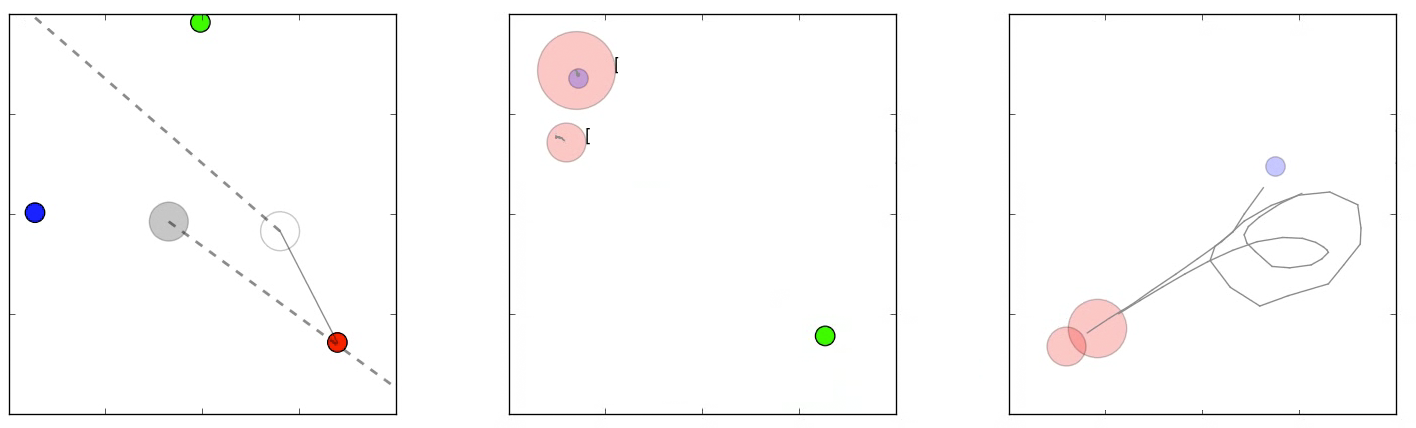}{0.45}{Examples of non-verbal communication strategies, such as pointing, guiding, and pushing.}

\section{Conclusion}
\label{sec:Conclusion}

We have presented a multi-agent environment and learning methods that brings about emergence of an abstract compositional language from grounded experience. This abstract language is formed without any exposure to human language use. We investigated how variation in environment configuration and physical capabilities of agents affect the communication strategies that arise.

In the future, we would like experiment with larger number of actions that necessitate more complex syntax and larger vocabularies. We would also like integrate exposure to human language to form communication strategies that are compatible with human use. 


\section*{Acknowledgements}
We thank OpenAI team for helpful comments and fruitful discussions. This work was funded in part by ONR PECASE N000141612723.

\bibliographystyle{aaai}
\bibliography{multiagent}

\section*{Appendix: Physical State and Dynamics}
\label{sec:appendix}

The physical state of the agent is specified by $\mathbf{x} = \begin{bmatrix} \ \mathbf{p} \ \dot{\mathbf{p}} \ \mathbf{v} \ \mathbf{d} \ \end{bmatrix}$ where $\dot{\mathbf{p}}$ is the velocity of $\mathbf{p}$. $\mathbf{d} \in \mathcal{R}^3$ is the color associted with the agent. Landmarks have similar state, but without gaze and velocity components. The physical state transition dynamics for a single agent $i$ are given by:
\begin{align*}\label{eq:dynamics}
\mathbf{x}_i^t = 
    \begin{bmatrix}
        \mathbf{p} \\
        \dot{\mathbf{p}} \\
        \mathbf{v}
    \end{bmatrix}_i^t = 
    \begin{bmatrix}
        \mathbf{p} + \dot{\mathbf{p}} \Delta t \\
        \gamma \dot{\mathbf{p}} + (\mathbf{u}_p + \mathbf{f}(\mathbf{x}_1,...,\mathbf{x}_N)) \Delta t \\
        \mathbf{u}_v \\
    \end{bmatrix}_i^{t-1}
\end{align*}

Where $\mathbf{f}(\mathbf{x}_1,...,\mathbf{x}_N)$ are the physical interaction forces (such as collision) between all agents in the environment and any obstacles, $\Delta t$ is the simulation timestep (we use 0.1), and $(1-\gamma)$ is a damping coefficient (we use 0.5). The action space of the agent is $\mathbf{a} = \begin{bmatrix} \ \mathbf{u}_p \ \mathbf{u}_v \ \mathbf{c} \ \end{bmatrix}$. The observation of any location $\mathbf{p}_j$ in reference frame of agent $i$ is $_i\mathbf{p}_j = R_i (\mathbf{p}_j - \mathbf{p}_i)$, where $R_i$ is the random rotation matrix of agent $i$. Giving each agent a private random orientation prevents identifying landmarks in a shared coordinate frame (using words such as \emph{top-most} or \emph{left-most}).


\end{document}